\documentclass[11pt,a4paper]{article}
\usepackage[hyperref]{acl2021}
\usepackage{times}
\usepackage{latexsym}

\usepackage{microtype}
\usepackage{makecell}
\usepackage{multirow}
\usepackage{amsmath}
\usepackage{booktabs}       
\usepackage{xargs}
\usepackage{graphicx}
\usepackage[colorinlistoftodos,prependcaption,textsize=small]{todonotes}

\usepackage{multirow}

\usepackage{paralist}
\usepackage{amsmath}

\newcommand{\Sref}[1]{\S\ref{#1}}

\usepackage{microtype}

\aclfinalcopy 

\title{Simple and Efficient ways to Improve REALM}

\author{Vidhisha Balachandran$^1$\hspace{0.5em} Ashish Vaswani$^2$\hspace{0.5em} Yulia Tsvetkov$^1$\hspace{0.5em}  Niki Parmar$^2$\\ 
  $^1$ Language Technologies Institute, Carnegie Mellon University \\
  $^2$ Google Research \\
  \texttt{\{vbalacha,ytsvetko\}@cs.cmu.edu} \\ \texttt{\{avaswani, nikip\}@google.com} \\}

\date{}

\begin{document}
\maketitle
\begin{abstract}
Dense retrieval has been shown to be effective for retrieving relevant documents for Open Domain QA, surpassing popular sparse retrieval methods like BM25. REALM \cite{guu2020realm} is an end-to-end dense retrieval system that relies on MLM based pretraining for improved downstream QA efficiency across multiple datasets. We study the finetuning of REALM on various QA tasks and explore the limits of various hyperparameter and supervision choices. We find that REALM was significantly undertrained when finetuning and simple improvements in the training, supervision, and inference setups can significantly benefit QA results and exceed the performance of other models published post it. 
Our best model, REALM++, incorporates all the best working findings and achieves significant QA accuracy improvements over baselines ($\sim$5.5\% absolute accuracy) without any model design changes. Additionally, REALM++ matches the performance of large Open Domain QA models which have 3x more parameters demonstrating the efficiency of the setup.

\end{abstract}

\section{Introduction}
Open-domain question answering (QA) \cite{voorhees1999trec} is a task that aims to answer questions directly using a large collection of documents without the context of a specific document. These systems generally employ a ``retriever-reader'' based approach where a \emph{document retriever} first retrieves a subset of evidence documents and a \emph{document reader} processes the documents to identify the correct answer \cite{chen2017reading}. Recently, dense retrieval methods \cite{seo2018phrase, seo2019real, guu2020realm, karpukhin2020dense} have made training these systems end-to-end by performing a search based semantic similarity complementary to sparse retrievers like BM25 \cite{Robertson2009ThePR}. 
REALM \cite{guu2020realm} is an end-to-end general retriever, pre-trained on masked language modeling that can be finetuned for downstream QA applications without relying on external sources like BM25 for supervision. 

In this paper, we present a study of REALM aimed at understanding the limitations of the system. We find that REALM is significantly under-optimized when finetuning on downstream QA tasks and is not directly comparable to models published after it as they use greater resources for better optimization. We show that simple improvements in the training, supervision, and inference setups when finetuning REALM for downstream QA tasks can significantly improve end QA accuracy and establish stronger performance.

We improve the training of REALM by scaling the system through (i) using exact MIPS search, (ii) introducing larger batch training, and (iii) scaling the reader to process more documents. 
We further address the limitations of potentially noisy distant supervision by augmenting the training sets with human-annotated evidence passages. Additionally, since obtaining such human annotations is expensive and is not available for every dataset, we show that models trained with strong supervision transfer well to other datasets where such annotations are not available, indicating the benefits beyond a single annotated dataset.
Finally, we experiment with scaling our document reader to process 10x more documents during inference. 

We call our system REALM++ and show that our simple improvements are highly efficient and outperform all prior methods which are of similar parameter regime across three Open Domain QA benchmarks.
Our results demonstrate that training and supervision choices play an important role in training Open Domain QA systems and that simple and straightforward improvements to the experimental setup of REALM can significantly benefit the system making it competitive with other recently published methods.

\section{REALM}
Below we briefly describe the REALM model and experimental setup and present a study on the performance of REALM to identify potential bottlenecks.

\paragraph{Background: }Open Domain QA is typically modeled as a Machine Reading Comprehension (MRC) model $M$ which takes question $Q$ to predict an answer $A$ using a large corpus $D$ of $m$ text documents/passages $D = \{D_1, D_2, ... D_m\}$. REALM \cite{guu2020realm, lee2019latent} specifically uses dense retrieval by learning document representations $d = \{h_{d_1}, h_{d_2}, ... h_{d_m}\}$ and question representation $h_{q}$ and using approximate MIPS to retrieve a large set of top $c$ relevant passages $P = \{p_1, ...., p_c\}$ with their dense representations $\{h_{p_1}, ...., h_{p_c}\}$. A smaller set of top $k$ passages based on the retrieval score, $S_{\mathit{retr}}(p_i, q) = h_q ^\top h_{p_i}$, are passed to the transformer based reader module (e.g BERT) to process the tokens in the question and passages to find the answer as a span in the passage. When finetuning, the retriever is trained using distant supervision considering any passage that contains the target answer as positive. The reader is supervised by predicting the target answer span for each passage. The retriever and reader are jointly trained using the distant passage supervision and strong span level short answer supervision. 

For our experiments, we follow the same setup of finetuning REALM on QA with the same optimization and evaluation setup. REALM retrieves $c=5000$ documents for distant passage supervision and $k=5$ documents are processed by the reader for finding the target answer. Similar to REALM, we use pre-trained checkpoints based on masked language modeling on the CC-News corpus. While REALM uses TPUs for distributed training for pretraining on MLM objective, the system is finetuned on a single machine with a 12GB GPU with batch size 1. 

\paragraph{Datasets:}
For our study, we use three open-domain QA datasets following \citet{guu2020realm}:\\
\emph{Natural Questions (NQ)} \cite{kwiatkowski2019natural} contains real user queries from Google Search. We consider questions with short answers ($<$=5 tokens) and the long answers for passage supervision.\\
\emph{WebQuestions (WQ)} \cite{berant2013semantic} is a
collection of questions collected from Google Suggest API, with Freebase entity answers whose string forms are the target short answers. \\
\emph{CuratedTREC (CT)} \cite{baudivs2015modeling} contains curated questions from TREC QA track with real user questions and answer as regular expression matching all acceptable answers.\\

\begin{table}[t]
    \setlength\tabcolsep{5pt}
    \centering
    \begin{tabular}{lccccc} \toprule
    {Metric} & \thead{NQ} & \thead{WQ} & \thead{CT}  \\ \midrule
    Test EM \cite{guu2020realm} & 40.4 & 40.7 & 42.9 \\
    Test EM & 39.4 & 40.8 & 39.3 \\
    \midrule
    Dev EM & 35.6&	45.4&	42.9\\
    Dev EM Upper Bound &63.3&	75.9&	70.8\\
    \midrule
    R@5 & 63.9&	78.5&	70.0\\
    R@10 & 69.4&	85.7&	76.4\\
    R@100 & 80.7&	94.0&	85.6\\
    R@1000 & 86.8&	97.7&	91.7\\
    R@5000 & 91.1&	99.2&	93.3\\
    \bottomrule
    \end{tabular}
    \caption{
    \textbf{Experiments reproduce results of REALM on NQ dataset}. First section compares Test EM from our experiments with previous published results from \cite{guu2020realm}. The middle section compares Dev EM with Upper Bound performance of the Reader and the bottom section compares answer recall at various subsets of retrieved documents.
    }
    \label{tab:qa_realm}
\vspace{-0.4cm}
\end{table}

\paragraph{Experiments fairly reproduce results: } Table \ref{tab:qa_realm} reports the results from our experiments and compares them to published results from REALM \cite{guu2020realm}. We find that our experiments produce similar results on NQ and WQ with slightly lower results on CT on the test set. We believe that this could be due to varying checkpoints due to early stopping. For fair evaluation, we use results from our experiments as a comparison for the remainder of the study. 

\paragraph{Answer recall drops significantly with reducing documents: }We additionally present a breakdown of REALM's retriever and reader performances on the development set across the three datasets. While REALM retrieves $c=5000$  documents for distantly supervising the retriever, only the top $k=5$ documents are processed by the reader for finding the right answer. Comparing the recall of answers in the retrieved documents at different subsets of documents we observe very high ($>90\%$) recall@5000 for all three datasets but the recall@5 effectively drops to $\sim$70\%, showing that the document that contains the answer is not necessarily present in the top-5 highlighting limitations in the retriever. 

\paragraph{Wide margins exist for improving reader performance: }Comparing the Exact Match accuracy of the system with the upper bound (the system is right if the passage contains the answer) shows wide gaps in the performance of the reader. While $\sim$63\% of the questions from NQ have the answer in the top retrieved documents,  REALM is only able to get the exact span of the answer for $\sim$36\% of them showing the limitations of the reader in identifying the exact answer span in the document.

\section{Exploring limits of REALM}
\label{sec:expl}
To address some of the bottlenecks in REALM shown above, we introduce simple changes in the model setup to explore and study the limits of REALM. For this study, we focus on the NQ dataset. We begin by scaling the resources and experimenting with larger batch size training along with replacing approximate search with an exact search for retrieval in \Sref{sec:train}. We then experiment with improving the passage supervision for retrieval by leveraging human annotations from NQ in \Sref{sec:sup}. To improve the recall of answers in the documents processed by the reader, we experiment with introducing a simple transformer based reranker in \Sref{sec:reranker}. Finally, we experiment with increasing the documents processed by the reader during inference in \Sref{sec:inf} to make the system comparable to prior methods.

\subsection{Scaling the Training Setup } 
\label{sec:train}
When finetuning for QA, REALM performs an approximate MIPS search for retrieving relevant documents. Additionally, the system is trained on a single machine with a 12GB GPU with $k=5$ and batch size 1. While this is modest use of resources, we show that this results in suboptimal training of the system. We begin by scaling the realm system during training. We perform exact MIPS search by leveraging the efficiency of large matrix multiplications of TPUs \cite{wang2019benchmarking} to compute the retrieval score, $S_{retr}$, for $\sim$13M passages of corpus and extract $c$ passages having the highest scores. We further increase the training batch size from $1$ to $16$ by leveraging 8 TPUv3 cores on Google Cloud for distributed training. Finally, we increase the number of documents passed to the reader to $k=10$ during training.

\paragraph{Scaling training setup improves QA results: } The results from improving the training of REALM through purely experimental scaling improvements are presented in Table \ref{tab:qa_abl}. We observe that such simple improvements to the experimental setup like larger batch training and exact MIPS search significantly improve the test-Acc by $3.4\%$ without introducing any model design changes. This shows that the original REALM setup was under-optimized and has much better performance than previously reported.

\begin{table}[t]
    \setlength\tabcolsep{5pt}
    \centering
    \begin{tabular}{lccccc} \toprule
    {Experiments} & \thead{Test\\Acc} & \thead{Dev\\Acc} & \thead{Dev\\R@10}  \\ 
    \midrule
    REALM & 39.4 & 35.6 &  68.8    \\
    +Scale & 42.8 & 37.9 & 69.5    \\
    +Scale+PS (10 docs inf) & 43.2 & 38.6 & 69.9  \\
    +Scale+PS (100 docs inf)& 44.8 & 38.6 & 69.9   \\
    +Scale+Rerank & 42.3 & 37.4 & 67.5   \\
    \bottomrule
    \end{tabular}
    \caption{
    \textbf{Scaling during training, Passage supervision and Inference document scaling improve Test EM Acc.} Test indicates test set, Dev indicated development set
    }
    \label{tab:qa_abl}
\vspace{-0.4cm}
\end{table}

\begin{table*}[h]
    \small
    \begin{tabular}{ p{2.5cm} p{5.2cm}  p{7cm} }
        \toprule
\textbf{Question}
& \textbf{Incorrect Passage}
& \textbf{Correct Passage from Human Annotations} \\\midrule
Where did the idea of a unicorn come from?
    & Unicorn is a privately held startup company whose name was coined in 2013 by venture capitalist Aileen Lee.
    & Unicorns are not found in Greek mythology, but rather in the accounts of natural history, for Greek writers of natural history were convinced of the reality of unicorns. 
    \\\hline
What type of reproduction do whiptail lizards use?          
    & MLB Whiparound is an American baseball television show on Fox Sports 1 hosted by Chris Myers and Kevin Burkhardt.
    & The New Mexico whiptail lizard is a crossbreed of a western whiptail and the little striped whiptail. The lizard is a female-only species that reproduces asexually by producing an egg through parthenogenesis.
    \\\hline
Which president supported the creation of the Environmental Protection Agency(EPA)? 
    & Some historians say that President Richard Nixon's southern strategy turned the southern United States into a republican stronghold, while others deem economic factors more important in the change.
    & The Environmental Protection Agency (EPA) is an agency of the federal government of the United States created for the purpose of protecting human health and the environment. President Richard Nixon proposed the establishment of EPA and it began operation on December 2, 1970, after Nixon signed an executive order.
    \\\bottomrule
    \end{tabular}
    \caption{Examples of Questions from Natural Questions with incorrect retrieved passages from \citet{guu2020realm} with the correct human annotated relevant passages showing the necessity for human annotation based supervision.}
    \label{tab:framework}
\vspace{-0.3cm}
\end{table*}

\subsection{Introducing Strong Passage Supervision }
\label{sec:sup}
For training the retriever, REALM relies on distant supervision in the form of passages containing the target answer since they are applicable to multiple datasets \cite{lee2019latent}. However, such a signal can lead to noisy and unrelated documents to be given a positive signal \cite{lin2018denoising}.
Table \ref{tab:framework} shows more examples of noisy passages from distant supervision. 

We address this by introducing strong supervision from human annotations. Our passage supervision considers the larger set of $c$ retrieved passages and uses human annotations to update the retrieval scores by optimizing their marginal log-likelihood .
\vspace{-0.4cm}
\begin{align*}
P(p_i | Q) &= \frac{\exp(S_{\mathit{retr}}(p_i, Q))}{\displaystyle\sum_{p_j \in {p_1, p_2...., p_c}}\exp(S_{\mathit{retr}}(p_j, Q))}\\
L(Q, LA) &= -\log\hspace{-0.5cm}\sum_{\substack{p_i \in {p_1, p_2...., p_c}\\ p_i \in LA}} \hspace{-0.5cm} P(p_i | Q)
\end{align*}

where $LA$ is a list of human annotations of evidence passages (e.g. Long Answers in Natural Questions), $L(Q, LA)$ denotes the passage supervision loss that is augmented to the existing retriever distant supervision and span prediction loss, and $p_i \in LA$ indicates whether the passage was in the annotated passages. Here, we assume that both passages in corpus $D$ and the annotated passages in the dataset are from the same source (e.g. Wikipedia). Since corpus passages and annotated passages in the dataset can differ (e.g. due to different Wikipedia versions), we consider any passage in the retrieved set that has 50\%\footnote{We experimented with different thresholds (0.3, 0.5, 0.75) and used the threshold with best performance based on validation set} word overlap\footnote{We also experimented with ngram overlap which did not improve performance significantly. We decided to use the word overlap metric as it is cheaper to compute.} with the target passages as a positive match.

\paragraph{Passage Supervision through Long Answer an-notations improves performance: }
Table \ref{tab:qa_abl} shows the results of using passage supervision using annotated evidence passages from the NQ dataset. We see an improvement of $3.8\%$ over the baseline REALM model showing the benefit of providing strong supervision for passage retrieval. Augmenting passage supervision to the scaled REALM model improves test Accuracy by $0.5$\% showing that strong supervision for passage relevance provides a good signal for training. In \Sref{sec:qual} we present qualitative examples of how better passage supervision leads to answer spans being extracted from the right document relevant to the question. Our experiments show that distant supervision can provide ambiguous and noisy supervision and annotated evidence passages can help with providing clean supervision.

\begin{table*}[t]
    \setlength\tabcolsep{5pt}
    \centering
    \begin{tabular}{lccccc} \toprule
    {Model} & {R@10} & {R@100} & {R@5000} & {Dev EM}  \\ 
    \midrule
    REALM & 68.8 & 81.5 & 91.1 & 35.6   \\
    \midrule
    +Scale (Fixed Ret) & 59.6 & 73.5 & 84.3 & 33.1   \\
    +Scale + Rerank (Fixed Ret) & 67.9 & 77.2 & 84.3 & 35.8  &   \\
    +Scale + Rerank + PS (Fixed Ret) & 67.5 & 76.5 & 84.3 & 37.1  &   \\ \midrule
    +Scale (Trained Ret) & 69.5 & 78.9 & 86.0 & 37.9 &   \\
    +Scale + Rerank (Trained Ret) & 67.5 & 78.9 & 86.0 & 37.4 &   \\
    \bottomrule
    \end{tabular}
    \caption{
    Comparing reranking based results for fixed v/s finetuned retriever. \textbf{Reranking significantly improves recall and dev EM in the when the retriever is fixed} but does not improve over scaled baseline when the retriever is trained.
    }
    \label{tab:qa_fixret}
\end{table*}

\subsection{Reranking}
\label{sec:reranker}
Table \ref{tab:qa_realm} shows that though recall of answer is high in the retrieved 5000 documents, the recall falls significantly when considering only the top 5 or 10 documents processed by the reader. Since readers are generally computationally intensive and scaling them to process many documents is not always feasible, we explore a simple approach to rerank the 5000 documents post retrieval to improve recall@10 and end accuracy. 

Our Document Reranker consists of $L$ layers of cross-document and query-document attentions to learn rich document representations for reranking. 
For each layer, the output passage representations from the previous layer,
$u^{l-1} = \{u^{l-1}_1, u^{l-1}_2, ... u^{l-1}_c\}$, are first passed through a Transformer block (Transf) with multi-headed self-attention \cite{vaswani2017attention} which allows for interaction between the passage representations. Every document $u^{l-1}_i$ attends to all other documents $\{u^{l-1}_1..u^{l-1}_c\}$ to produce cross-document aware representations $v^{l-1}_i$.
\begin{multline}
    u^{l}_i = \text{Transf}(Q=u^{l-1}_i, K=\{u^{l-1}_1..u^{l-1}_c\},\\ V=\{u^{l-1}_1..u^{l-1}_c\})
\end{multline}
where $Q$ represents the query, $K$ represents the key and $V$ represents the value in the transformer attention module. To model interaction between passages and the query, the attended passage representations and the query representation from the previous layer $v^{l-1}$ are passed through a multi-head cross-attention Transformer where question representation $v^{l-1}$ attends to the attended document representations $\{u^{l}_1..u^{l}_c\}$.
\begin{multline}
    v^{l}_q = \text{Transf}(Q=v^{l-1}_q, K=\{u^{l}_1..u^{l}_c\},\\ V=\{u^{l}_1..u^{l}_c\})
\end{multline}
For the first layer we consider retrieved document representations $u^{0} = \{h_1, h_2, ... h_c\}$ and query representation $v^0_q = h_q$ from retriever as the input. The rich document representations and query representations from the final layer, $\{u^{L}_1..u^{L}_c\}$ and $v^L_q$ are used to compute the retriever score, $S_{\mathit{retr}}(p_i, q)$ for extracting the top-k documents for the reader.

\paragraph{Document Reranking does not provide significant gains when the retriever is jointly trained: }Table \ref{tab:qa_abl} shows the results of augmenting the above document reranker with our scaled REALM model. We observe that the accuracy of the system drops by $0.5\%$ and the recall@10 drops by $0.2\%$. While this suggests that the reranking module did not contribute to improving the end-to-end trained system, we further study the role of the reranker in a fixed retriever setting where the top 5000 documents are retrieved once and kept constant during training. 

\paragraph{Document Reranking is highly effective when retriever is fixed: }While such a setting is a more efficient and faster version of the jointly trained model as relevant documents are not retrieved at every training step, the model's zero-shot performance of the retriever can be quite low, potentially hurting end accuracy. From Table \ref{tab:qa_fixret}, we see that the REALM model when trained with a fixed-retriever, has very low recall@Top-10 and development set Exact-Match. Here, augmenting the model with the Document Reranker significantly improves recall and EM performance, where recall@Top-10 improves by $8.3\%$ and EM by $2.7\%$. Further introducing passage supervision during training improves performance by increasing the end accuracy by $\sim1.3\%$ making the fixed retriever setting very competitive in performance to a jointly trained retriever-reader setting using a more efficient and faster setup.

\begin{table*}[t]
    \setlength\tabcolsep{5pt}
    \centering
    \begin{tabular}{lccccc} \toprule
    {Model} & \thead{NQ (79k/4k)}  & \thead{WQ (3k/2k)} & \thead{CT (1k/1k)}  \\ \midrule
    BM25+BERT~\cite{lee2019latent} & 26.5  & 17.7 & 21.3  \\
    {ORQA}~\cite{lee2019latent} & 33.3  & 36.4 & 30.1    \\
    HardEM~\cite{min2019discrete} & 28.1  & - & -  \\
    GraphRetriever~\cite{min2019knowledge} & 34.5  & 36.4 & -  \\
    PathRetriever~\cite{asai2020learning}  & 32.6  & - & - \\
    {REALM}~\cite{guu2020realm} & 39.2  & 40.2 & 46.8  \\ 
    {REALM}$_\textrm{News}$~\cite{guu2020realm} & 40.4  & 40.7 & 42.9  \\ 
    {DPR}~\cite{karpukhin2020dense} & 41.5 & 42.4 & 49.4  \\ 
    {ReConsider}$_\textrm{Base}$~\cite{iyer2020reconsider} & 43.1 & 44.4 & 49.3  \\ 
    \midrule
    REALM++ (10 docs) & 43.2 &  44.5$^*$  & 47.2$^*$  \\
    REALM++ (100 docs) & \textbf{44.8} &  \textbf{45.6}$^*$  & \textbf{49.7}$^*$  \\
    \midrule
    {DPR-BERT}$_\textrm{Large}$~\cite{iyer2020reconsider} & 44.6 & 44.8 & 53.5  \\ 
    {RAG}$_\textrm{Large}$~\cite{lewis2020retrievalaugmented} & 44.5 & 45.5 & 52.2  \\ 
    {ReConsider}$_\textrm{Large}$~\cite{iyer2020reconsider} & \textbf{45.5} & \textbf{45.9} & \textbf{55.3} \\ 
    \bottomrule
    \end{tabular}
    \caption{
    Test QA (Exact Match) Accuracy on Open-QA benchmarks showing \textbf{REALM++ improving over prior methods of similar size.} The number of train/test examples are shown in parentheses next to each benchmark. *indicates models finetuned on trained NQ model. Top section show performance of Open Domain QA systems which use base models and bottom section show performance of systems which use large models.
    }
    \label{tab:qa_em}
\end{table*}


\subsection{Scaling the Reader at Inference}
\label{sec:inf}
Due to accelerator memory constraints, the reader can only process a limited set of documents during training $k=5,10$ and it cannot be scaled to process more documents directly without architectural changes. We experiment with addressing this limitation by scaling the reader to process more documents during inference. During inference, since optimization based weights and parameters are not saved in memory, the memory usage of the model reduces, allowing for the reader to process more documents. 
\begin{figure}[t]
    {\includegraphics[width=\columnwidth]{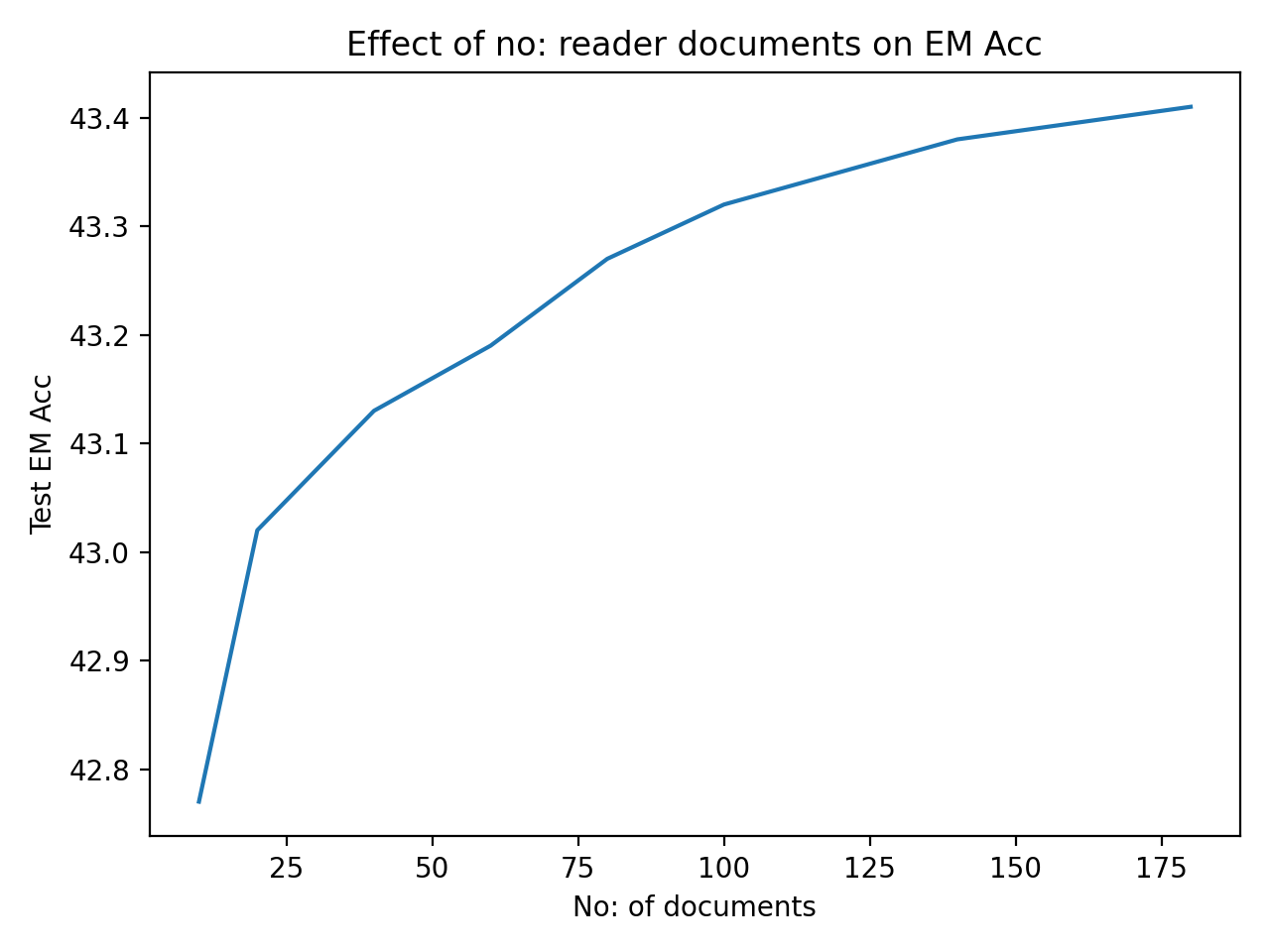}}
    \caption{\textbf{Test QA Accuracy of Scaled REALM increases with increasing documents processed by the reader during inference.} No: of documents are increased from 10 to 180.}
    \label{fig:inf_res}
\end{figure}

\paragraph{Scaling the reader during inference significantly boosts performance: } In Table \ref{tab:qa_abl} we see that extending the reader to process $k=100$ documents significantly improves end QA accuracy, achieving $44.8\%$ accuracy on NQ which surpasses the baseline REALM by $4.4\%$. This shows that REALM based QA systems can leverage a small number of documents ($k=10$) for faster training and gain the benefits of scaling the reader to a large number of documents ($k=100$) at inference. 

Further from Figure \ref{fig:inf_res} we see that this improvement increases with increasing number documents processed by the reader incrementally from $k=10$ to $k=180$ with a slight saturation beyond 120 documents. The end QA accuracy increases potentially due to increased answer recall among more documents. We see the trend increasing even beyond 100 documents, but we choose 100 documents for reporting and for our best model to make the model more comparable to other methods which use a similar number of documents for training and inference. An interesting point to note is that increasing the documents processed by the reader can increase inference speeds and presents an interesting speed v/s accuracy trade-off for practitioners. We show in Table \ref{tab:qa_em} in \Sref{sec:realm+} that our results outperform other state-of-art methods on 2 out of 3 datasets even when using $k=10$ documents, demonstrating that our setup is better even when prioritizing speed.

\section{REALM ++}
\label{sec:realm+}
Based on the findings from our experiments in \Sref{sec:expl}, we incorporate the best working components: (i) scaling during training \Sref{sec:train} (ii) strong passage supervision \Sref{sec:sup} and (iii) scaling reader during inference \Sref{sec:inf} to establish an improved REALM model, which we call REALM++. We study the effect of REALM++ on the QA accuracy on three datasets NQ, WQ, and CT. As WQ and CT do not have evidence passage annotations, we use them to study the transfer capabilities of the passage supervised NQ model similar to DPR. We compare REALM++ to prior extractive (the answer is a span in the document) and generative (the answer is generated) Open Domain QA models to understand how our simple improvements fare against other systems.

\begin{table*}[h]
    \small
    \begin{tabular}{ p{2.5cm} p{6cm}  p{6cm} }
        \toprule
\textbf{Question}
& \textbf{Incorrect Ret Passage}
& \textbf{Correct REALM++ Ret Passage} \\\midrule
Where did the Battle of Issus take place?
    & The Battle of Alexander at Issus is a 1529 oil painting by the German artist Albrecht Altdorfer, a pioneer of landscape art and a founding member of the Danube School .
    & The Battle of Issus occurred in southern Anatolia, on November 5, 333 BC between the Hellenic League led by Alexander the Great and the Achaemenid Empire, led by Darius III. 
    \\\hline
Who played bubba in the Heat of the Night?
    & A late Stevan Ridley touchdown run set up by a 23 - yard Deangelo Peterson run on a fourth - down play gave LSU the upset victory and effectively ended the opportunity for an Alabama repeat of the national championship.
    & Carlos Alan Autry Jr. is an American actor, politician, and former National Football League player. He played the role of Captain Bubba Skinner on the NBC television series, "In the Heat of the Night", starring Carroll O'Connor.
    \\\hline
Actress who plays penelope garcia on criminal minds?
    & How to get away with Murder is an American television series created by Peter Nowalk and produced by Shonda Rhimes and ABC Studios.
    & Kirsten Simone Vangsness is an American actress, currently staring as FBI Analyst Penelope Garcia on the CBS series "Criminal Minds".
    \\\bottomrule
    \end{tabular}
    \caption{Qualitative Analysis of questions from NQ showing questions where baseline REALM retrieved incorrect passages and \textbf{training with passage supervision helped retrieve the right passage}.}
    \label{tab:qual_ex}
    \vspace{-0.3cm}
\end{table*}

\paragraph{REALM++ outperforms prior models of similar size and is comparable to large models:}
Table \ref{tab:qa_em} shows that while REALM++ significantly improves over the baseline REALM model by $4.4$\% on the NQ dataset, it additionally outperforms prior methods of similar size (models based on BERT$_{base}$) with no modifications to the model design. When transferred to WQ and CT which do not have human annotation for passage supervision, REALM++ shows an improvement of $3.8$\% on WQ and $4.3$\% on CT over base REALM. This shows that while sourcing human annotations for evidence passages can be expensive, their benefit can be transferred well to other datasets increasing their use beyond a single dataset. REALM++ produces state-of-art results on extractive Open Domain QA among models of similar size in all three datasets using a single end-to-end model.

From Table \ref{tab:qa_em} we observe that our REALM++ model which uses BERT$_{base}$ ($\sim$ 110M parameters) performs comparable to large models based on BERT$_{large}$ and BART$_{large}$ \cite{lewis-etal-2020-bart} ($\sim$ 340M parameters) with 3x lesser parameters. While ReConsider$_{large}$ \cite{iyer2020reconsider} has higher accuracy numbers, their approach of using an additional model for answer span focused reranking is orthogonal to our model and can be directly applied to our outputs.

\paragraph{Discussion of speed and memory usage: }
By using 8 TPUv3 cores and increased batch size for training our REALM++ model, we can process 4x more examples/sec as compared to REALM and reduce training time from 2 days to 12hrs. 
Further, as we train for the same number of epochs, similar improvements in accuracy could potentially be obtained using distributed training on multiple GPUs or gradient accumulation on a single GPU.
Exact top-K increases the memory utilization by 5GB since we store the entire document index in memory. Our entire model fits within 12GB memory which is the equivalent of an Nvidia Titan X. 
This demonstrates that our REALM++ model is efficient and can improve training time by leveraging distributed training.

\subsection{Qualitative Analysis}
\label{sec:qual}
In \Sref{sec:sup}, we introduced strong passage supervision from annotated evidence passages to enable the model to distinguish misleading passages that might contain the target answer. Table \ref{tab:qual_ex} shows examples of questions where using passage supervision helps retrieve correct passages for the QA task. For Questions 1 and 3, the baseline model incorrectly retrieves a wrong passage of a similar genre or topic as the question, while for Question 2 the baseline model retrieves a completely incorrect, irrelevant passage. The model trained with passage supervision identifies the right context for answering the question, which aligns with the human annotation for each question. 

\section{Related Work}
\paragraph{Open Domain QA:} Open Domain QA \cite{voorhees1999trec} aims to answers questions directly based on a large corpus of unstructured documents. \citet{chen2017reading} use heuristic TF-IDF based retrievers for identifying relevant documents from a corpus and and RNN based reader for identifying the right answer within the document. Early systems used sparse retrieval like BM25 or TF-IDF for retrieving passages/documents \cite{chen2017reading, yang2019end, nie2019revealing, min2019knowledge}. \citet{min2019discrete, asai2020learning} explore augmenting sparse retrieval with structured knowledge via connecting graphs to improve passage retrieval. \citet{seo2018phrase, Das2019MultistepRI, lee2019latent, seo2019real} explore dense embedding based methods for retrieving relevant passage using semantic match between question and passage. DPR \cite{karpukhin2020dense} revisits pipelined QA systems where independent retrieval and reader systems are trained. The retrieval is a dense retrieval system trained using distant supervision from BM25. \citet{iyer2020reconsider} extend DPR by performing answer focused reranking. We focus on REALM \cite{guu2020realm}, which is an end-to-end dense-retrieval system, for the purpose of our study and reranking is orthogonal and can be applied to our system as well. \citet{lewis2020retrieval, roberts2020much} use an alternative approach to generate the answer instead of extracting it from the document. While this reduces the constraints on the model, generated answers are not grounded to the document creating potential for hallucinations. In contrast, REALM is constrained in ensuring the answer is always a span in the document reducing any hallucinatory effects. 

\paragraph{Passage Supervision:} Passage retrieval has been an important component for open-domain QA. \citet{Das2019MultistepRI, lin2018denoising, wang2018r} use distant supervision for training the retriever by considering a paragraph that contains the ground truth answer as a positive example. \citet{karpukhin2020dense} use a different form of distant supervision considering top retrieved BM25 passages as positive examples. Datasets like WikiQA \cite{yang-etal-2015-wikiqa} and MSMarco \cite{nguyen2016ms} provide gold supervision of candidate sentences or passages relevant to the question for supervision. Our system leverages similar candidate passage annotations from Natural Questions dataset for supervising the dense passage retrieval. While distant supervision has been shown to be effective for dense retrieval, our work experimentally shows the benefit of gold candidate passage supervision for training dense retrieval systems.


\section{Conclusion}
In this work, we present a study of a dense-retrieval QA system, REALM, and identify key limitations in its experimental setup. We find that REALM is significantly undertrained and we improve REALM by introducing simple changes to its training, supervision, and inference setup. We show that replacing approximate MIPS with exact search and using larger batch training can improve the base model establishing a stronger model. Further strong supervision through annotations of evidence passages can significantly improve the system outperforming prior models. While obtaining such annotations can be expensive, we show that models trained with such passage supervision transfer well to other datasets which do not have such human annotations making them generalizable beyond a single dataset. Finally, we show that scaling the reader during inference significantly boosts performance. We propose REALM++ which incorporates our best findings and show that it outperforms other Open Domain QA models.

\section*{Acknowledgments}
We thank Kalpesh Krishna and Aurko Roy for weekly discussions and feedback provided during the course of this work. We thank Kenton Lee, Kelvin Guu, Zora Tung and Ming-Wei Chang for discussions about REALM and help with codebase and Ruiqi Guo for help with efficient TPU top-K implementations. We are grateful to Bhuwan Dhingra and William Cohen for useful discussions and Dheeraj Rajagopal for help with reviewing the paper and providing feedback.

\bibliographystyle{acl_natbib}
\bibliography{anthology,acl2021}

\begin{thebibliography}{28}
\expandafter\ifx\csname natexlab\endcsname\relax\def\natexlab#1{#1}\fi

\bibitem[{Asai et~al.(2020)Asai, Hashimoto, Hajishirzi, Socher, and
  Xiong}]{asai2020learning}
Akari Asai, Kazuma Hashimoto, Hannaneh Hajishirzi, Richard Socher, and Caiming
  Xiong. 2020.
\newblock Learning to retrieve reasoning paths over {Wikipedia} graph for
  question answering.
\newblock In \emph{ICLR}.

\bibitem[{Baudi{\v{s}} and {\v{S}}ediv{\`y}(2015)}]{baudivs2015modeling}
Petr Baudi{\v{s}} and Jan {\v{S}}ediv{\`y}. 2015.
\newblock Modeling of the question answering task in the yodaqa system.
\newblock In \emph{International Conference of the Cross-Language Evaluation
  Forum for European Languages}.

\bibitem[{Berant et~al.(2013)Berant, Chou, Frostig, and
  Liang}]{berant2013semantic}
Jonathan Berant, Andrew Chou, Roy Frostig, and Percy Liang. 2013.
\newblock Semantic parsing on {Freebase} from question-answer pairs.
\newblock In \emph{EMNLP}.

\bibitem[{Chen et~al.(2017)Chen, Fisch, Weston, and Bordes}]{chen2017reading}
Danqi Chen, Adam Fisch, Jason Weston, and Antoine Bordes. 2017.
\newblock Reading {Wikipedia} to answer open-domain questions.
\newblock In \emph{ACL}.

\bibitem[{Das et~al.(2019)Das, Dhuliawala, Zaheer, and
  McCallum}]{Das2019MultistepRI}
Rajarshi Das, S.~Dhuliawala, M.~Zaheer, and A.~McCallum. 2019.
\newblock Multi-step retriever-reader interaction for scalable open-domain
  question answering.
\newblock \emph{ICLR}.

\bibitem[{Guu et~al.(2020)Guu, Lee, Tung, Pasupat, and Chang}]{guu2020realm}
Kelvin Guu, Kenton Lee, Zora Tung, Panupong Pasupat, and Ming-Wei Chang. 2020.
\newblock {REALM}: Retrieval-augmented language model pre-training.
\newblock In \emph{ICML}.

\bibitem[{Iyer et~al.(2020)Iyer, Min, Mehdad, and Yih}]{iyer2020reconsider}
Srinivasan Iyer, Sewon Min, Yashar Mehdad, and Wen-tau Yih. 2020.
\newblock Reconsider: Re-ranking using span-focused cross-attention for open
  domain question answering.
\newblock \emph{arXiv preprint arXiv:2010.10757}.

\bibitem[{Karpukhin et~al.(2020)Karpukhin, O{\u{g}}uz, Min, Lewis, Wu, Edunov,
  Chen, and Yih}]{karpukhin2020dense}
Vladimir Karpukhin, Barlas O{\u{g}}uz, Sewon Min, Patrick Lewis, Ledell Wu,
  Sergey Edunov, Danqi Chen, and Wen-tau Yih. 2020.
\newblock Dense passage retrieval for open-domain question answering.
\newblock In \emph{EMNLP}.

\bibitem[{Kwiatkowski et~al.(2019)Kwiatkowski, Palomaki, Redfield, Collins,
  Parikh, Alberti, Epstein, Polosukhin, Kelcey, Devlin, Lee, Toutanova, Jones,
  Chang, Dai, Uszkoreit, Le, and Petrov}]{kwiatkowski2019natural}
Tom Kwiatkowski, Jennimaria Palomaki, Olivia Redfield, Michael Collins, Ankur
  Parikh, Chris Alberti, Danielle Epstein, Illia Polosukhin, Matthew Kelcey,
  Jacob Devlin, Kenton Lee, Kristina~N. Toutanova, Llion Jones, Ming-Wei Chang,
  Andrew Dai, Jakob Uszkoreit, Quoc Le, and Slav Petrov. 2019.
\newblock Natural questions: a benchmark for question answering research.
\newblock \emph{TACL}.

\bibitem[{Lee et~al.(2019)Lee, Chang, and Toutanova}]{lee2019latent}
Kenton Lee, Ming-Wei Chang, and Kristina Toutanova. 2019.
\newblock Latent retrieval for weakly supervised open domain question
  answering.
\newblock In \emph{ACL}.

\bibitem[{Lewis et~al.(2020{\natexlab{a}})Lewis, Liu, Goyal, Ghazvininejad,
  Mohamed, Levy, Stoyanov, and Zettlemoyer}]{lewis-etal-2020-bart}
Mike Lewis, Yinhan Liu, Naman Goyal, Marjan Ghazvininejad, Abdelrahman Mohamed,
  Omer Levy, Veselin Stoyanov, and Luke Zettlemoyer. 2020{\natexlab{a}}.
\newblock \href {https://doi.org/10.18653/v1/2020.acl-main.703} {{BART}:
  Denoising sequence-to-sequence pre-training for natural language generation,
  translation, and comprehension}.
\newblock In \emph{Proceedings of the 58th Annual Meeting of the Association
  for Computational Linguistics}, pages 7871--7880, Online. Association for
  Computational Linguistics.

\bibitem[{Lewis et~al.(2020{\natexlab{b}})Lewis, Perez, Piktus, Petroni,
  Karpukhin, Goyal, Kuttler, Lewis, tau Yih, Rockt{\"a}schel, Riedel, and
  Kiela}]{lewis2020retrieval}
Patrick Lewis, Ethan Perez, Aleksandara Piktus, F.~Petroni, V.~Karpukhin, Naman
  Goyal, Heinrich Kuttler, M.~Lewis, Wen tau Yih, Tim Rockt{\"a}schel,
  Sebastian Riedel, and Douwe Kiela. 2020{\natexlab{b}}.
\newblock Retrieval-augmented generation for knowledge-intensive nlp tasks.
\newblock \emph{ArXiv}, abs/2005.11401.

\bibitem[{Lewis et~al.(2020{\natexlab{c}})Lewis, Perez, Piktus, Petroni,
  Karpukhin, Goyal, Küttler, Lewis, tau Yih, Rocktäschel, Riedel, and
  Kiela}]{lewis2020retrievalaugmented}
Patrick Lewis, Ethan Perez, Aleksandara Piktus, Fabio Petroni, Vladimir
  Karpukhin, Naman Goyal, Heinrich Küttler, Mike Lewis, Wen tau Yih, Tim
  Rocktäschel, Sebastian Riedel, and Douwe Kiela. 2020{\natexlab{c}}.
\newblock Retrieval-augmented generation for knowledge-intensive nlp tasks.
\newblock In \emph{neurips}.

\bibitem[{Lin et~al.(2018)Lin, Ji, Liu, and Sun}]{lin2018denoising}
Yankai Lin, Haozhe Ji, Zhiyuan Liu, and Maosong Sun. 2018.
\newblock Denoising distantly supervised open-domain question answering.
\newblock In \emph{Proceedings of the 56th Annual Meeting of the Association
  for Computational Linguistics (Volume 1: Long Papers)}, pages 1736--1745.

\bibitem[{Min et~al.(2019{\natexlab{a}})Min, Chen, Hajishirzi, and
  Zettlemoyer}]{min2019discrete}
Sewon Min, Danqi Chen, Hannaneh Hajishirzi, and Luke Zettlemoyer.
  2019{\natexlab{a}}.
\newblock A discrete hard {EM} approach for weakly supervised question
  answering.
\newblock In \emph{EMNLP}.

\bibitem[{Min et~al.(2019{\natexlab{b}})Min, Chen, Zettlemoyer, and
  Hajishirzi}]{min2019knowledge}
Sewon Min, Danqi Chen, Luke Zettlemoyer, and Hannaneh Hajishirzi.
  2019{\natexlab{b}}.
\newblock Knowledge guided text retrieval and reading for open domain question
  answering.
\newblock \emph{arXiv preprint arXiv:1911.03868}.

\bibitem[{Nguyen et~al.(2016)Nguyen, Rosenberg, Song, Gao, Tiwary, Majumder,
  and Deng}]{nguyen2016ms}
Tri Nguyen, Mir Rosenberg, Xia Song, Jianfeng Gao, Saurabh Tiwary, Rangan
  Majumder, and Li~Deng. 2016.
\newblock Ms marco: A human-generated machine reading comprehension dataset.

\bibitem[{Nie et~al.(2019)Nie, Wang, and Bansal}]{nie2019revealing}
Yixin Nie, Songhe Wang, and Mohit Bansal. 2019.
\newblock Revealing the importance of semantic retrieval for machine reading at
  scale.

\bibitem[{Roberts et~al.(2020)Roberts, Raffel, and Shazeer}]{roberts2020much}
Adam Roberts, Colin Raffel, and Noam Shazeer. 2020.
\newblock How much knowledge can you pack into the parameters of a language
  model?
\newblock \emph{EMNLP}.

\bibitem[{Robertson and Zaragoza(2009)}]{Robertson2009ThePR}
S.~Robertson and H.~Zaragoza. 2009.
\newblock The probabilistic relevance framework: Bm25 and beyond.
\newblock \emph{Found. Trends Inf. Retr.}, 3:333--389.

\bibitem[{Seo et~al.(2018)Seo, Kwiatkowski, Parikh, Farhadi, and
  Hajishirzi}]{seo2018phrase}
Minjoon Seo, T.~Kwiatkowski, Ankur~P. Parikh, Ali Farhadi, and Hannaneh
  Hajishirzi. 2018.
\newblock Phrase-indexed question answering: A new challenge for scalable
  document comprehension.
\newblock In \emph{EMNLP}.

\bibitem[{Seo et~al.(2019)Seo, Lee, Kwiatkowski, Parikh, Farhadi, and
  Hajishirzi}]{seo2019real}
Minjoon Seo, Jinhyuk Lee, Tom Kwiatkowski, Ankur Parikh, Ali Farhadi, and
  Hannaneh Hajishirzi. 2019.
\newblock Real-time open-domain question answering with dense-sparse phrase
  index.
\newblock In \emph{ACL}.

\bibitem[{Vaswani et~al.(2017)Vaswani, Shazeer, Parmar, Uszkoreit, Jones,
  Gomez, Kaiser, and Polosukhin}]{vaswani2017attention}
Ashish Vaswani, Noam Shazeer, Niki Parmar, Jakob Uszkoreit, Llion Jones,
  Aidan~N Gomez, Lukasz Kaiser, and Illia Polosukhin. 2017.
\newblock Attention is all you need.
\newblock In \emph{NIPS}.

\bibitem[{Voorhees et~al.(1999)}]{voorhees1999trec}
Ellen~M Voorhees et~al. 1999.
\newblock The {TREC-8} question answering track report.
\newblock In \emph{TREC}.

\bibitem[{Wang et~al.(2018)Wang, Yu, Guo, Wang, Klinger, Zhang, Chang, Tesauro,
  Zhou, and Jiang}]{wang2018r}
Shuohang Wang, Mo~Yu, Xiaoxiao Guo, Zhiguo Wang, Tim Klinger, Wei Zhang, Shiyu
  Chang, Gerry Tesauro, Bowen Zhou, and Jing Jiang. 2018.
\newblock R 3: Reinforced ranker-reader for open-domain question answering.
\newblock In \emph{Proceedings of the AAAI Conference on Artificial
  Intelligence}, volume~32.

\bibitem[{Wang et~al.(2019)Wang, Wei, and Brooks}]{wang2019benchmarking}
Yu~Emma Wang, Gu-Yeon Wei, and David Brooks. 2019.
\newblock Benchmarking tpu, gpu, and cpu platforms for deep learning.
\newblock \emph{arXiv preprint arXiv:1907.10701}.

\bibitem[{Yang et~al.(2019)Yang, Xie, Lin, Li, Tan, Xiong, Li, and
  Lin}]{yang2019end}
Wei Yang, Yuqing Xie, Aileen Lin, Xingyu Li, Luchen Tan, Kun Xiong, Ming Li,
  and Jimmy Lin. 2019.
\newblock End-to-end open-domain question answering with bertserini.
\newblock In \emph{HLT-NAACL}, pages 72--77.

\bibitem[{Yang et~al.(2015)Yang, Yih, and Meek}]{yang-etal-2015-wikiqa}
Yi~Yang, Wen-tau Yih, and Christopher Meek. 2015.
\newblock \href {https://doi.org/10.18653/v1/D15-1237} {{W}iki{QA}: A challenge
  dataset for open-domain question answering}.
\newblock In \emph{Proceedings of the 2015 Conference on Empirical Methods in
  Natural Language Processing}, pages 2013--2018, Lisbon, Portugal. Association
  for Computational Linguistics.

\end{thebibliography}


\end{document}